\documentclass{article}

\usepackage{arxiv}

\usepackage[utf8]{inputenc} 
\usepackage[T1]{fontenc}    
\usepackage{hyperref}       
\usepackage{url}            
\usepackage{booktabs}       
\usepackage{amsfonts}       
\usepackage{nicefrac}       
\usepackage{microtype}      
\usepackage{lipsum}
\usepackage{graphicx}
\usepackage{orcidlink}
\graphicspath{ {./images/} }

\title{Human-in-the-Loop Atlas-Based 3D Asset Segmentation for Interactive Content Workflows}

\author{
Paul Julius Kühn\textsuperscript{*} \\
Fraunhofer IGD, 64283 Darmstadt, Germany \And
Saptarshi Neil Sinha\textsuperscript{*} \\
Fraunhofer IGD, 64283 Darmstadt, Germany \And
Jakob Hansen \\
Fraunhofer IGD, 64283, Darmstadt Germany \And
Robin Horst \\
Fraunhofer IGD \& Hochschule RheinMain, Germany
}

\begin{document}
\maketitle
\renewcommand{\thefootnote}{\fnsymbol{footnote}}
\footnotetext[1]{Equal contribution}

Segmenting 3D assets into meaningful regions remains challenging, especially when segmentation criteria are application-dependent and require user control. We present a human-in-the-loop pipeline for generating a segmented 2D parameterized atlas from a 3D model for interactive media, game, and XR content workflows. Our method first selects a compact set of rendered views using a greedy set cover strategy over sampled surface points, and then supports interactive segmentation of these views with SAM~2 and Label Studio. The resulting masks are back-projected onto the model's UV parameterization to produce a unified segmented atlas that supports downstream production tasks such as segment-wise material assignment, style transfer, and semantic labeling. We assess the pipeline through a demonstration-based technical evaluation on eight cultural heritage objects. The results show that the approach can generate usable segmented atlases across diverse geometries while revealing recurring sources of manual correction, particularly fine structures, cavities, and weak appearance boundaries. The code is available at \url{https://github.com/saptarshineil/ai_assisted_atlas_segmentation}.
\section{Introduction}
\label{sec:introduction}

Segmenting 3D assets into meaningful regions remains a challenging problem~\cite{3d_segmentation_survey}. Existing 3D segmentation methods are often dataset-specific~\cite{dataset_specific}, requiring retraining for new object categories, or rely on primitive-based decompositions that assume simple geometric shapes~\cite{primitive_based_1,fedele2025superdec} and therefore struggle with complex real-world surfaces. More recent approaches~\cite{yang2024sampart3d} transfer 2D foundation model features to 3D backbones for zero-shot part segmentation, but they still rely on automatic semantic labeling and offer limited support for user-defined or application-specific segment boundaries. This is a practical limitation for interactive media, game, and XR content workflows, where segmentation is often needed not as an end in itself, but as a controllable intermediate representation for material editing, stylization, semantic labeling, or other texture-space operations.

In contrast, 2D image segmentation has advanced rapidly through foundation models such as SAM~2~\cite{ravi2024sam2}. A natural strategy is therefore to segment rendered views in 2D and project the results back onto the 3D surface. However, fully automatic use of such models remains insufficient in many scenarios. Even strong 2D foundation models can produce incomplete masks, over-segmentation, or incorrect boundaries, and the notion of a meaningful segment often depends on the intended application. In cultural heritage, segmentation criteria may evolve as new conservation or material information becomes available; in other domains, users may require different region boundaries for different downstream tasks. A human-in-the-loop workflow is therefore needed, in which users retain control over segmentation semantics while AI supports the process through editable mask predictions.

Atlas-based segmentation through multi-view 2D renderings is a promising strategy for controllable 3D asset processing, as it enables users to operate in image space while transferring the results back to the surface parameterization. However, existing atlas-based workflows still leave two key challenges insufficiently addressed: viewpoint selection is often not explicitly optimized for coverage, and segmentation is frequently treated as a fully automatic prediction problem without interactive correction or refinement. To address these limitations, we introduce a path-planning-inspired view selection stage and a human-in-the-loop segmentation workflow that supports iterative refinement of segmentation masks across rendered views. In light of these improvements, our main contributions are:
\begin{itemize}
    \item A \textbf{path-planning-inspired view selection algorithm} that formulates viewpoint planning as a set cover problem and uses a greedy strategy to obtain a compact set of camera positions covering the sampled surface representation.
    \item A \textbf{complete end-to-end human-in-the-loop pipeline} integrating SAM~2 and Label Studio~\cite{label_studio} for iterative refinement of segmentation predictions across views, producing a segmented 2D parameterized atlas for downstream tasks such as segment-wise shading, material assignment, and other texture-space editing workflows.
    \item A \textbf{demonstration-based evaluation} on eight cultural heritage objects, characterizing annotation effort, output adequacy, and recurring failure modes across diverse geometries and appearance conditions.
\end{itemize}

\section{Related works}
\label{sec:related_works}

\begin{figure*}[htb!]
  \centering
  \includegraphics[width=\textwidth]{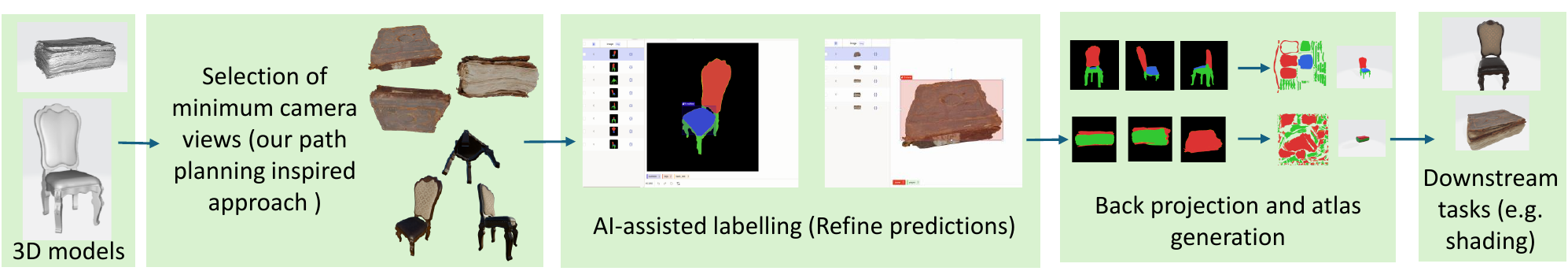}
  \caption{Overview of the proposed pipeline for generating a segmented 2D parameterized atlas from a 3D model: Given an input 3D model, a minimal set of camera views is first selected using our path-planning-inspired approach to ensure full surface coverage. AI-assisted labelling is then performed on the rendered views, with iterative refinement of multi-view predictions to ensure cross-view consistency. The segmented labels are back-projected onto the model's UV parameterization to produce a fully segmented 2D atlas, which can subsequently be used for downstream applications such as segment-wise shading, style transfer, etc.}
  \label{fig:pipeline_full}
\end{figure*}

Segmenting 3D models into meaningful parts is a long-standing challenge in computer graphics and geometry processing~\cite{3d_segmentation_survey}. Early methods relied on geometric criteria to partition meshes into perceptually meaningful components. Katz et al.~\cite{katz2003hierarchical} proposed a hierarchical decomposition algorithm that segments meshes at regions of deep concavities using fuzzy clustering and graph cuts. Lavoué et al.~\cite{lavoue2005new} introduced a segmentation method based on curvature tensor field analysis, decomposing surfaces into near-constant curvature patches with clean boundaries. Shapira et al.~\cite{shapira2008consistent} developed the shape diameter function, a volume-based measure that enables consistent partitioning across families of objects in different poses and resolutions. Liu and Zhang~\cite{liu2004segmentation} formulated mesh segmentation as a spectral clustering problem on face affinity matrices, favoring cuts along concave regions inspired by human perception. While these geometric approaches produce meaningful results, they typically require careful parameter tuning and remain tied to specific object categories~\cite{chang2015shapenet}.

A separate family of 3D segmentation methods decomposes surfaces into simple geometric primitives. Schnabel et al.~\cite{schnabel2007efficient} presented an efficient RANSAC-based algorithm that automatically detects planes, spheres, cylinders, cones, and tori in unorganized point clouds, decomposing them into a concise hybrid structure of inherent shapes and remaining points. Region growing approaches~\cite{rabbani2006segmentation} similarly fit local primitive models to progressively expand coherent surface patches. More recently, SuperDec~\cite{fedele2025superdec} employs a Transformer-based architecture to decompose point clouds into superquadric primitives, achieving compact yet expressive scene representations that generalize from object-level training to full 3D scenes. While primitive-based methods are effective for CAD models and architectural environments composed of regular shapes, they fundamentally assume that surfaces conform to a limited vocabulary of analytic forms and struggle with complex, organic, or irregular real-world objects. 

Recently, deep learning has significantly advanced 3D segmentation. PointNet~\cite{qi2016pointnet} pioneered direct consumption of unordered point sets for classification and segmentation tasks. PointNet++~\cite{qi2017pointnetplusplus} extended this with hierarchical feature learning that captures local geometric structures at increasing contextual scales. Wang et al.~\cite{wang2019dynamic} proposed Dynamic Graph CNN, which constructs graphs dynamically in each network layer to recover topology and capture semantic characteristics over long distances. Despite their effectiveness, these supervised methods require large-scale annotated 3D datasets and do not generalize to unseen categories without retraining. The success of 2D foundation models such as SAM~2~\cite{ravi2024sam2} has inspired efforts to transfer their segmentation capabilities to 3D. SA3D~\cite{cen2023segment} extends SAM to Neural Radiance Fields by alternating mask inverse rendering and cross-view self-prompting, achieving 3D segmentation from a single-view prompt within minutes. PartSLIP~\cite{liu2023partslip} transfers knowledge from the pretrained image-language model GLIP~\cite{li2021grounded} to 3D point clouds through part detection on rendered views and a novel 2D-to-3D label lifting algorithm, enabling zero-shot and few-shot 3D part segmentation. SAMPart3D~\cite{yang2024sampart3d} distills text-agnostic vision foundation model features into a 3D backbone trained on large-scale datasets, achieving scalable zero-shot part segmentation at multiple granularities. However, all these methods rely on automatic semantic labeling with no mechanism for users to define or correct what constitutes a meaningful segment for their specific application.

An alternative strategy projects 2D segmentation results from multiple rendered views back onto a 3D model's surface parameterization. Prior work~\cite{our_previous_paper} demonstrated the feasibility of this atlas-based approach using Tracking-Anything-with-DEVA~\cite{cheng2023tracking} as a fully automatic video segmentation backbone, enabling downstream applications such as semantic stylization. However, that approach employed a fixed camera animation without explicit viewpoint optimization for coverage and offered no mechanism for human correction or refinement. In contrast, our approach combines path-planning-inspired viewpoint selection with a human-in-the-loop workflow that integrates SAM~2~\cite{ravi2024sam2} and an interactive annotation platform to support user-guided refinement of segmentation results across views.

\section{Methodology}
\label{sec:methodology}

\begin{figure*}[htb!]
  \centering
  \includegraphics[width=\textwidth]{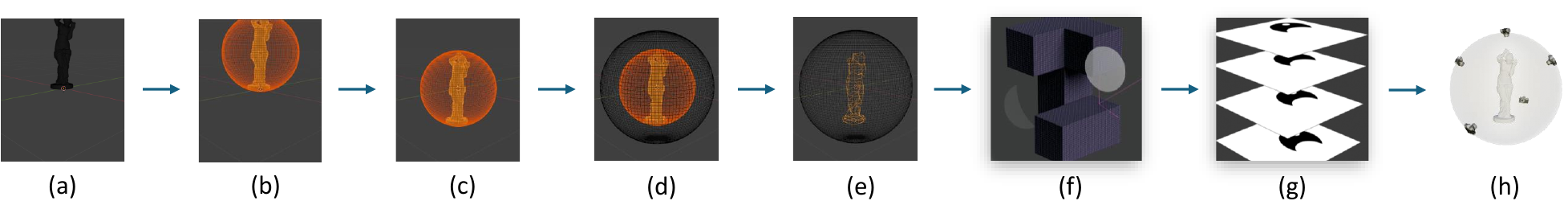}
  \caption{Pipeline for selecting a minimal set of camera views for full surface coverage of a 3D model: (a) Load the model; (b) Compute bounding sphere; (c) Translate sphere and object to origin; (d) Compute camera sphere; (e) Remesh model for performance; (f) Build visibility maps using raycasts from each surface point to each camera position; (g) Iteratively select the camera position with the highest surface coverage and remove associated visibility map from stack; (h) Final output with minimum set of camera views.}
  \label{fig:path_planning}
\end{figure*}

  \begin{figure*}[htb!]
  \centering
  \includegraphics[width=\textwidth]{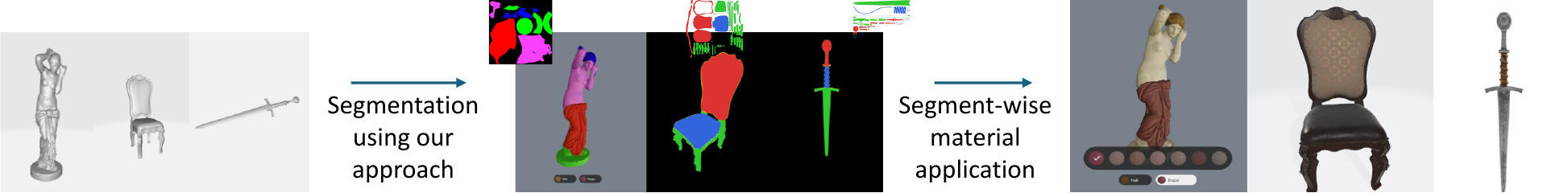}
  \caption{Example downstream tasks of segment-wise material application: Three objects (a statue, a Victorian chair, and a sword) are first segmented using our approach and then shaded with distinct materials assigned per segment. The resulting textured models are visualized in both WebGL and OpenGL viewers, demonstrating the practical applicability of our segmented 2D parameterized atlas for interactive rendering workflows.}
  \label{fig:example_shading_our_approach}
\end{figure*}

This section presents our pipeline for generating a fully segmented 2D parameterized atlas from a 3D model (see Figure~\ref{fig:pipeline_full}). The approach consists of four stages: (A) selection of a minimal set of camera views that guarantee full surface coverage using our path-planning-inspired approach, (B) AI-assisted labelling of the rendered views with iterative refinement of predictions, (C) back-projection of the segmented colour images onto the UV atlas, and (D) downstream tasks enabled by the resulting segmented atlas, such as segment-wise shading.

The pipeline assumes as input a 3D model equipped with a 2D parameterized atlas (UV map) that contains no overlapping regions, as overlaps would introduce ambiguity during the back-projection stage where image pixels are mapped to unique texel locations. If the model does not already possess a suitable atlas, one can be generated as a preprocessing step using, for example, Blender's Smart UV Project functionality. However, the user must ensure that the resulting UV islands do not overlap, as this is not guaranteed by default. With a valid non-overlapping atlas in place, the pipeline proceeds as follows.
\subsection{Selection of minimum camera views}
The first step in the pipeline (see Figure~\ref{fig:pipeline_full}) is to obtain the minimum set of camera positions that ensure every point on the 2D parameterized surface of the 3D model is visible from at least one viewpoint.  Taking inspiration from path planning techniques used in robotics, such as viewpoint planning for robotic arm inspection tasks, we formulate this as a set cover problem over the model's surface. The algorithm (see Figure~\ref{fig:path_planning}), implemented within BlenderProc~\cite{blenderProc}, begins by loading the 3D model and enclosing it in a computed bounding sphere, then translating both so their shared center lies at the origin (see Figure~\ref{fig:path_planning}(a)-(c)). A camera sphere is constructed at a fixed radius from the origin, defining the complete set of candidate viewpoints (see Figure~\ref{fig:path_planning}(d)). The model is optionally remeshed to produce a uniform distribution of sample points across its surface parameterization, enabling consistent and efficient visibility evaluation (see Figure~\ref{fig:path_planning}(e)). Visibility maps are then constructed by raycasting from each surface sample point to every candidate camera position on the camera sphere, encoding which viewpoints can observe each surface point via equirectangular projection (see Figure~\ref{fig:path_planning}(f)). This computation is parallelized to manage cost. Camera positions are then selected iteratively using a greedy strategy: at each step, the candidate position from which the greatest number of previously uncovered surface points are visible is chosen, and those points are removed from further consideration (see Figure~\ref{fig:path_planning}(g)). The process repeats until every sampled point on the model's surface is covered by at least one camera position. Finally, the remeshed model is replaced with the original high-fidelity mesh, keyframes are set at the selected positions, metadata is recorded, and images are rendered from the chosen viewpoints (Fig.~\ref{fig:path_planning}(h)).

\subsection{AI assisted segmentation}
Once the minimal set of views has been rendered, semantic segmentation is performed on each image using an AI-assisted interactive annotation workflow. We employ Label Studio~\cite{label_studio} as the annotation platform, integrated with the SAM~2~\cite{ravi2024sam2} as the underlying foundation model for mask prediction. Foundation models such as SAM~2 often predict segments correctly out of the box, producing high-quality masks with minimal user input. However, a fully automated approach is insufficient in practice. Even when segmentation criteria remain fixed, SAM~2 can produce erroneous results such as incomplete masks, over-segmentation, or incorrect boundary delineation that require human correction. Moreover, segmentation criteria are not universal and may evolve over time depending on the application domain and available information. For instance, in remote sensing, soil region boundaries may shift with changing weather conditions and updated sensor readings; in cultural heritage, the criteria for segmenting an object may be revised as new historical, material, or conservation information becomes available through different sensing paradigms such as multispectral imaging or X-ray fluorescence. This inherent variability in what constitutes a meaningful segment motivates our human-in-the-loop design, where domain experts retain control over the segmentation semantics while leveraging the efficiency of foundation model predictions.

For each rendered view, the annotator provides minimal prompts (e.g., point clicks or bounding boxes) to indicate regions of interest, and SAM~2 generates segmentation masks for the indicated parts. This interactive setup significantly reduces manual effort compared to fully manual polygon annotation, while retaining human oversight to correct erroneous predictions and adapt the segmentation to domain-specific criteria. To ensure cross-view consistency, predictions are iteratively refined across the set of rendered views.  The final output of this stage is a set of per-view segmentation masks with consistent class labels, ready for back-projection onto the model's UV parameterization.
\subsection{Backprojection onto the 2D parameterized atlas}
With per-view segmented colour images obtained from the AI-assisted labelling stage, the next step is to transfer these colour-coded labels from image space onto the model's 2D parameterized atlas (UV map). For each selected camera position, the corresponding segmented colour image is projected back onto the 3D surface by establishing a correspondence between image pixels and mesh faces. For every pixel in the segmented image, we determine which mesh face is visible at that pixel using an index pass rendered from the same viewpoint, and then write the pixel's segment colour to the corresponding location in UV space using the face's texture coordinates.

This process is repeated for each of the selected camera views, producing a set of partial atlas images, each containing colour-coded segment information only for the surface regions visible from its respective viewpoint. Since the camera selection stage guarantees that every point on the model's surface is observed by at least one view, the union of all partial projections provides complete coverage of the UV atlas. The partial atlas images are then merged into a single, unified segmented atlas. The resulting merged atlas is a fully segmented 2D parameterization of the model's surface, where each texel carries a colour corresponding to its semantic class, ready for use in downstream applications.
\subsection{Downstream applications via the 2D parameterized atlas}
With a fully segmented 2D parameterized atlas in hand, a range of geometry and graphics tasks can be performed directly in UV space. Segment-wise material application becomes simple and unambiguous, as each region on the atlas can be independently assigned distinct reflectance model parameters such as roughness, metallicity, and albedo without manual masking (Fig.~\ref{fig:example_shading_our_approach}). Style transfer can likewise be applied on a per-region basis, leveraging existing 2D neural style transfer methods that operate directly on the unwrapped surface. Further applications include texture synthesis constrained to individual segments, automated 3D surface annotation for dataset generation, and semantic labelling for downstream machine learning tasks.
\begin{figure}[htb!]
  \centering
  \includegraphics[width=0.52\linewidth]{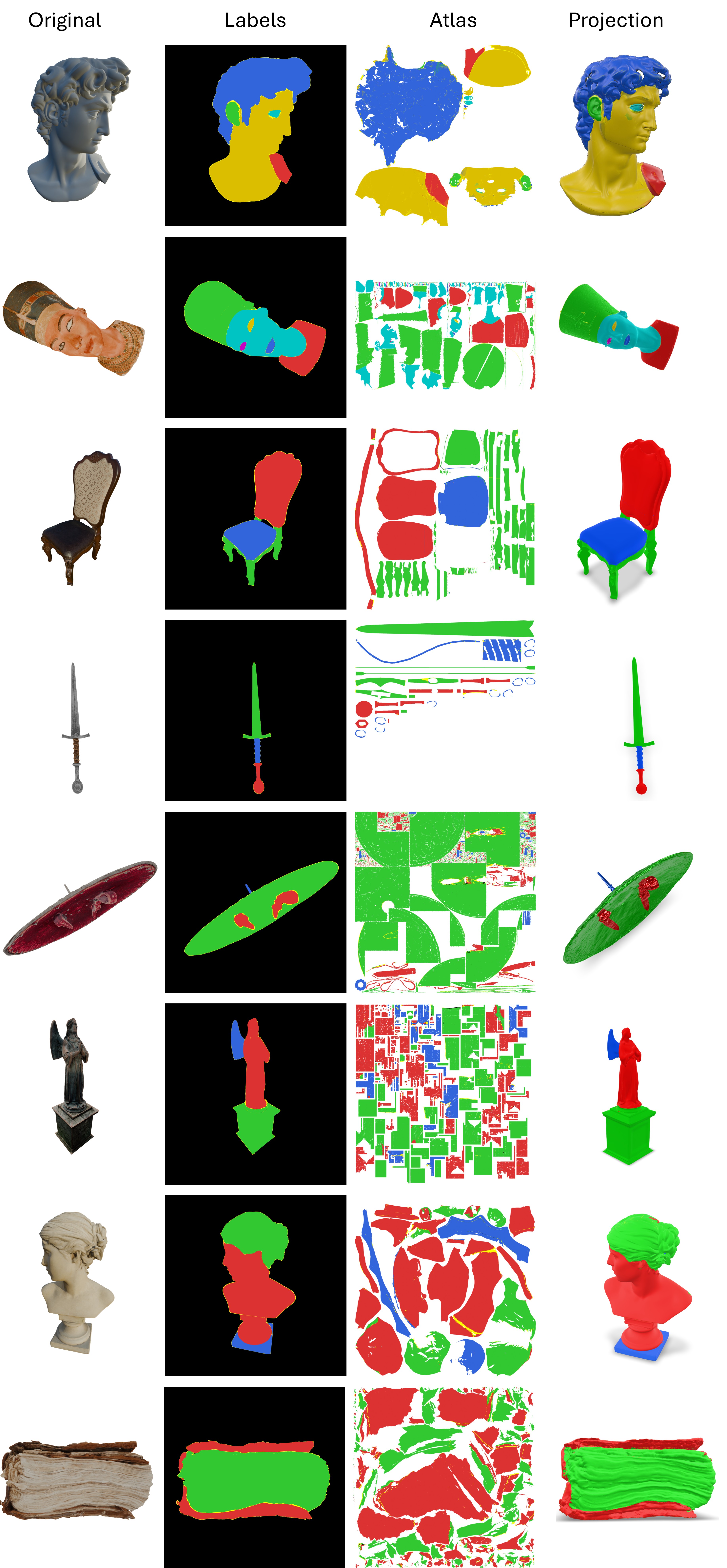}
  \caption{Evaluation examples top to bottom: Head of Michelangelo's David, a bust of Nefertiti, a Victorian chair, a medieval sword, the parade shield of King Erik XIV of Sweden, an angel statue, a bust of R\'o\.{z}a Loewenfeld, and a Coptic prayer book.}
  \label{fig:eval_examples}
\end{figure}

\section{Evaluation}

We evaluate the proposed pipeline through demonstration on representative problem instances conducted by the authors. Following a design science research perspective, this constitutes an ex post artificial evaluation of a technical artifact on realistic domain instances rather than a participant-based user study~\cite{peffers2007design,venable2016feds}. The goal is therefore to assess technical feasibility, output adequacy, and recurring failure modes, rather than user acceptance or deployment in professional practice.

We selected eight cultural heritage objects with substantially different geometry and appearance (Fig.~\ref{fig:eval_examples}): the head of Michelangelo's David, a bust of Nefertiti, a Victorian chair, a medieval sword, the parade shield of King Erik XIV of Sweden, an angel statue, a bust of R\'o\.{z}a Loewenfeld, and a Coptic prayer book. Although cultural heritage serves as the demonstration domain, the pipeline itself is not domain-specific. The objects were chosen to cover smooth marble surfaces, weak appearance boundaries, thin structures, cavities, repeated textures, and fine semantic details. 
Under our fixed configuration, the view-planning stage produced eight rendered views per model. The coverage objective is defined over the sampled surface points used during planning; accordingly, full coverage in this algorithmic sense does not preclude localized projection artifacts or weak viewing angles on the original high-fidelity mesh. The rendered views were segmented in Label Studio using SAM~2-assisted interaction and manually refined where necessary before being back-projected onto the UV atlas. We recorded approximate total annotation time per object, object-specific difficulties, and qualitative observations regarding coverage and correction effort (Fig.~\ref{tab:demonstration_objects}). In this evaluation, we consider an atlas usable if the relevant object regions can be delineated with acceptable manual correction effort and the resulting atlas can be employed for downstream segment-wise material assignment without major missing regions.

\begin{table*}[htb]
\centering
\caption{Objects used in the demonstration and main observations.}
\label{tab:demonstration_objects}
\scalebox{0.88}{
\begin{tabular}{|l|p{4.9cm}|p{1.2cm}|p{7.7cm}|}
\hline
\textbf{Object} & \textbf{Main challenge} & \textbf{Approx. time} & \textbf{Result summary} \\
\hline
Head of \emph{David} & Low texture/contrast; fine facial details & $\sim$15 min & Broad regions segmented reliably; eyes and ears required largely manual annotation \\
\hline
Nefertiti bust & Fine eye structures; ambiguous neck texture & $\sim$35 min & Good overall coverage, but the most time-consuming case due to correction effort around eyes and neck \\
\hline
Victorian chair & Underside cavity & $\sim$20 min & Good overall coverage; the underside cavity remained the most difficult region \\
\hline
Medieval sword & No major issues & $\sim$15 min & Fully covered with little correction effort \\
\hline
Parade shield & Back-side straps; narrow boundaries & $\sim$20 min & Good overall coverage; straps required manual boundary correction \\
\hline
Angel statue & Wings visually similar to body & $\sim$25 min & Complete coverage achieved; wings required manual drawing \\
\hline
R\'o\.{z}a Loewenfeld bust & Hollow interior; weak top-view coverage & $\sim$15 min & Usable overall result; localized artifact on the upper hair region \\
\hline
Coptic prayer book & Low contrast between cover and pages & not logged & Good overall coverage; simple global structure was advantageous \\
\hline
\end{tabular}%
}
\end{table*}

\subsection{Results}
Across all eight objects, the full pipeline could be executed successfully and yielded segmented atlases that were usable for our downstream material-assignment examples. For the seven objects for which annotation times were logged, total effort ranged from approximately 15 to 35 minutes, with a mean of about 21 minutes. Simpler objects such as the medieval sword required little correction, whereas busts and statues with weak appearance cues or small semantic regions required substantially more manual intervention.

\begin{figure}[htb!]
  \centering
  \includegraphics[width=.8\linewidth]{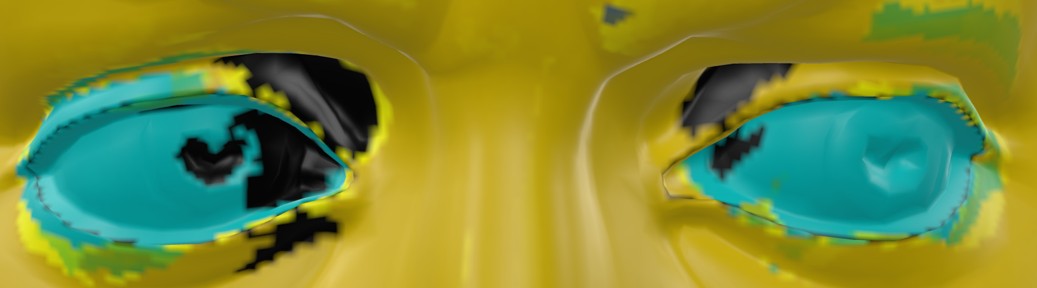}
  \caption{Segmentation artifacts in the projected eye region of the David statue. Fine eye structures required manual refinement. Black regions indicate localized projection artifacts caused by unfavorable viewpoints, while greenish regions show overlapping projected label colors.}
  \label{fig:eyes}
\end{figure}

The head of David was among the most challenging cases. As a marble object with almost no color variation and only weak texture cues, it offered limited guidance for SAM~2. In practice, explicit scene lighting was essential to create shading differences that made AI-assisted annotation useful. Larger regions such as hair, neck, and most of the face were segmented reliably, whereas eyes (Fig.~\ref{fig:eyes}) and ears had to be annotated largely by hand. A similar pattern appeared for the angel statue, where the wings were frequently merged with the main body because they shared nearly identical appearance. The Nefertiti bust likewise showed that fine structures around the eyes and ambiguities in the neck region can increase correction effort considerably, making it the most time-consuming object in the demonstration.

Objects with cavities or self-occlusion highlighted both the advantages and the limits of the multi-view strategy. For the Victorian chair, the underside cavity was less favorably covered than the outer surfaces, although the final atlas still remained largely complete. On the parade shield, the arm straps on the back side required manual boundary correction because both SAM~2 and Label Studio struggled with narrow structures. The bust of R\'o\.{z}a Loewenfeld further indicated that the current viewpoint selection can still produce locally unfavorable angles, leading to a visible artifact in the upper hair region.

The simplest cases were the medieval sword and the Coptic prayer book. For the sword, no substantial difficulties were observed and the atlas could be generated with minimal correction effort. The prayer book likewise benefited from a simple global structure, although the low contrast between cover and pages occasionally reduced the quality of automatic masks.

\subsection{Cross-case observations and limitations}
Several recurring patterns emerged across the demonstration. First, SAM~2 provided the greatest benefit on larger contiguous regions with sufficiently clear appearance or shading cues. Second, manual intervention remained necessary for very small structures, narrow attachments, and adjacent regions with nearly identical appearance. Third, the multi-view setup proved beneficial because multiple views could compensate for missing or imprecise boundaries in individual images, although shallow viewing angles still reduced precision in cavities and interior surfaces.

A further practical limitation was the annotation interface itself. While Label Studio supported efficient refinement in general, highly precise corrections for very small regions remained cumbersome. Thus, the bottleneck in difficult cases was not only the segmentation model, but also the granularity of interactive editing.

This evaluation has several methodological limitations. It is a demonstration-based technical assessment rather than a participant-based user study and therefore supports claims about feasibility, correction effort, and output characteristics, but not about user acceptance or professional adoption. The reported times are approximate and were not collected under controlled experimental conditions. No ground-truth 3D segmentations were available, so the evaluation focuses on practical adequacy rather than absolute accuracy. Finally, the selected objects serve as representative demonstration instances rather than as a basis for statistical generalization.

\section{Conclusion}
\label{sec:conclusion}
We presented a human-in-the-loop pipeline for atlas-based 3D segmentation that combines a greedy set-cover view selection algorithm guaranteeing complete surface coverage with minimal viewpoints, and an interactive SAM~2-based annotation workflow enabling expert-guided refinement of segmentation masks. The back-projected results yield a fully segmented 2D parameterized atlas suitable for downstream tasks such as segment-wise material assignment and style transfer. Our demonstration on eight cultural heritage objects showed that the pipeline can generate usable segmented atlases across diverse assets, with SAM~2 providing substantial assistance on larger contiguous regions, while fine details, low-contrast boundaries, and cavity-like structures remain the main sources of manual correction.

Future work includes conducting a larger-scale user study involving more domain experts across diverse application fields to enable a more robust statistical analysis of efficiency and segmentation quality. Another promising direction is leveraging the expert-annotated data produced through our pipeline to embed task-specific priors into the foundation model, progressively reducing human intervention and enabling fully automatic or minimal-labelling segmentation for well-defined, recurring tasks.
\section{Acknowledgments}
We thank Luis Scholl and Jannis Neus for their contributions to the implementation of the path planning approach.

\bibliographystyle{unsrt}  
\bibliography{references}  

\begin{thebibliography}{10}

\bibitem{3d_segmentation_survey}
Yong He, Hongshan Yu, Xiaoyan Liu, Zhengeng Yang, Wei Sun, Saeed Anwar, and Ajmal Mian.
\newblock Deep learning based 3d segmentation in computer vision: A survey.
\newblock {\em Information Fusion}, 115:102722, 2025.

\bibitem{dataset_specific}
Vladimir~G. Kim, Wilmot Li, Niloy~J. Mitra, Siddhartha Chaudhuri, Stephen DiVerdi, and Thomas Funkhouser.
\newblock Learning part-based templates from large collections of 3d shapes.
\newblock {\em ACM Trans. Graph.}, 2013.

\bibitem{primitive_based_1}
Florent Lafarge and Cl{\'e}ment Mallet.
\newblock Creating large-scale city models from 3d-point clouds: A robust approach with hybrid representation.
\newblock {\em International Journal of Computer Vision}, 2012.

\bibitem{fedele2025superdec}
Elisabetta Fedele, Boyang Sun, Leonidas Guibas, Marc Pollefeys, and Francis Engelmann.
\newblock {SuperDec: 3D Scene Decomposition with Superquadric Primitives}.
\newblock In {\em {Proceedings of the IEEE/CVF International Conference on Computer Vision (ICCV)}}, 2025.

\bibitem{yang2024sampart3d}
Yunhan Yang, Yukun Huang, Yuan-Chen Guo, Liangjun Lu, Xiaoyang Wu, Lam Edmund~Y., Yan-Pei Cao, and Xihui Liu.
\newblock Sampart3d: Segment any part in 3d objects.
\newblock {\em arXiv preprint arXiv:2411.07184}, 2024.

\bibitem{ravi2024sam2}
Nikhila Ravi, Valentin Gabeur, Yuan-Ting Hu, Ronghang Hu, Chaitanya Ryali, Tengyu Ma, Haitham Khedr, Roman R{\"a}dle, Chloe Rolland, Laura Gustafson, Eric Mintun, Junting Pan, Kalyan~Vasudev Alwala, Nicolas Carion, Chao-Yuan Wu, Ross Girshick, Piotr Doll{\'a}r, and Christoph Feichtenhofer.
\newblock Sam 2: Segment anything in images and videos.
\newblock {\em arXiv preprint arXiv:2408.00714}, 2024.

\bibitem{label_studio}
Maxim Tkachenko, Mikhail Malyuk, Andrey Holmanyuk, and Nikolai Liubimov.
\newblock {Label Studio}: Data labeling software, 2020-2025.
\newblock Open source software available from https://github.com/HumanSignal/label-studio.

\bibitem{katz2003hierarchical}
Sagi Katz and Ayellet Tal.
\newblock Hierarchical mesh decomposition using fuzzy clustering and cuts.
\newblock {\em ACM Trans. Graph.}, 2003.

\bibitem{lavoue2005new}
Guillaume Lavoué, Florent Dupont, and Atilla Baskurt.
\newblock A new cad mesh segmentation method, based on curvature tensor analysis.
\newblock {\em Computer-Aided Design}, 2005.

\bibitem{shapira2008consistent}
Lior Shapira, Ariel Shamir, and Daniel Cohen-Or.
\newblock Consistent mesh partitioning and skeletonisation using the shape diameter function.
\newblock {\em The Visual Computer}, 2008.

\bibitem{liu2004segmentation}
Rong Liu and Hao Zhang.
\newblock Segmentation of 3d meshes through spectral clustering.
\newblock In {\em 12th Pacific Conference on Computer Graphics and Applications, 2004. PG 2004. Proceedings.}, 2004.

\bibitem{chang2015shapenet}
Angel~X. Chang, Thomas Funkhouser, Leonidas Guibas, Pat Hanrahan, Qixing Huang, Zimo Li, Silvio Savarese, Manolis Savva, Shuran Song, Hao Su, Jianxiong Xiao, Li~Yi, and Fisher Yu.
\newblock Shapenet: An information-rich 3d model repository, 2015.

\bibitem{schnabel2007efficient}
Ruwen Schnabel, Roland Wahl, and Reinhard Klein.
\newblock Efficient {RANSAC} for point-cloud shape detection.
\newblock {\em Comput. Graph. Forum}, 2007.

\bibitem{rabbani2006segmentation}
T~Rabbani~Shah, FA~van~den Heuvel, and MG~Vosselman.
\newblock Segmentation of point clouds using smoothness constraint.
\newblock In H-G Maas and D~Schneider, editors, {\em Proceedings of the ISPRS Com. V Symposium}, pages 248--253. Dresden University of Technology, 2006.

\bibitem{qi2016pointnet}
Charles~R Qi, Hao Su, Kaichun Mo, and Leonidas~J Guibas.
\newblock Pointnet: Deep learning on point sets for 3d classification and segmentation.
\newblock {\em arXiv preprint arXiv:1612.00593}, 2016.

\bibitem{qi2017pointnetplusplus}
Charles~R Qi, Li~Yi, Hao Su, and Leonidas~J Guibas.
\newblock Pointnet++: Deep hierarchical feature learning on point sets in a metric space.
\newblock {\em arXiv preprint arXiv:1706.02413}, 2017.

\bibitem{wang2019dynamic}
Yue Wang, Yongbin Sun, Ziwei Liu, Sanjay~E. Sarma, Michael~M. Bronstein, and Justin~M. Solomon.
\newblock Dynamic graph cnn for learning on point clouds.
\newblock {\em ACM Trans. Graph.}, 2019.

\bibitem{cen2023segment}
Jiazhong Cen, Zanwei Zhou, Jiemin Fang, Chen Yang, Wei Shen, Lingxi Xie, Dongsheng Jiang, Xiaopeng Zhang, and Qi~Tian.
\newblock Segment anything in 3d with nerfs.
\newblock In {\em Proceedings of the 37th International Conference on Neural Information Processing Systems}, 2023.

\bibitem{liu2023partslip}
Minghua Liu, Yinhao Zhu, Hong Cai, Shizhong Han, Zhan Ling, Fatih Porikli, and Hao Su.
\newblock Partslip: Low-shot part segmentation for 3d point clouds via pretrained image-language models.
\newblock In {\em Proceedings of the IEEE/CVF Conference on Computer Vision and Pattern Recognition (CVPR)}, 2023.

\bibitem{li2021grounded}
Liunian~Harold Li, Pengchuan Zhang, Haotian Zhang, Jianwei Yang, Chunyuan Li, Yiwu Zhong, Lijuan Wang, Lu~Yuan, Lei Zhang, Jenq-Neng Hwang, Kai-Wei Chang, and Jianfeng Gao.
\newblock Grounded language-image pre-training.
\newblock In {\em CVPR}, 2022.

\bibitem{our_previous_paper}
Saptarshi~Neil Sinha, Paul~Julius Kühn, Pavel Rojtberg, Holger Graf, Arjan Kuijper, and Michael Weinmann.
\newblock {Semantic Stylization and Shading via Segmentation Atlas utilizing Deep Learning Approaches}.
\newblock In {\em Smart Tools and Applications in Graphics - Eurographics Italian Chapter Conference}. The Eurographics Association, 2024.

\bibitem{cheng2023tracking}
Ho~Kei Cheng, Seoung~Wug Oh, Brian Price, Alexander Schwing, and Joon-Young Lee.
\newblock Tracking anything with decoupled video segmentation.
\newblock In {\em ICCV}, 2023.

\bibitem{blenderProc}
Maximilian Denninger, Dominik Winkelbauer, Martin Sundermeyer, Wout Boerdijk, Markus Knauer, Klaus~H. Strobl, Matthias Humt, and Rudolph Triebel.
\newblock Blenderproc2: A procedural pipeline for photorealistic rendering.
\newblock {\em Journal of Open Source Software}, 8(82):4901, 2023.

\bibitem{peffers2007design}
Ken Peffers, Tuure Tuunanen, Marcus~A Rothenberger, and Samir Chatterjee.
\newblock A design science research methodology for information systems research.
\newblock {\em Journal of management information systems}, 24(3):45--77, 2007.

\bibitem{venable2016feds}
John Venable, Jan Pries-Heje, and Richard Baskerville.
\newblock Feds: a framework for evaluation in design science research.
\newblock {\em European journal of information systems}, 25(1):77--89, 2016.

\end{thebibliography}

\end{document}